%% file: seal_arxiv.tex
\pdfoutput=1

\documentclass[11pt]{article}

\usepackage[final]{acl}

\usepackage{times}
\usepackage{latexsym}

\usepackage[T1]{fontenc}

\usepackage[utf8]{inputenc}

\usepackage{microtype}

\usepackage{inconsolata}

\usepackage{graphicx}

\input{math_commands.tex}

\usepackage{hyperref}
\usepackage{url}

\usepackage{graphicx}
\usepackage{tcolorbox}
\usepackage{kotex}

\usepackage[capitalize]{cleveref}

\usepackage{caption}
\usepackage{subcaption}
\usepackage{booktabs}
\usepackage{tabularx}
\usepackage{graphicx}
\usepackage{multirow}
\usepackage{multicol}
\crefname{section}{Sec.}{Secs.}
\Crefname{section}{Section}{Sections}
\Crefname{table}{Table}{Tables}
\crefname{table}{Tab.}{Tabs.}

\newcommand{\ie}{\textit{i}.\textit{e}., }
\newcommand{\eg}{\textit{e}.\textit{g}.\ }

\title{SEAL: Scaling to Emphasize Attention for Long-Context Retrieval}

\author{Changhun Lee$^{1}$ \quad Minsang Seok$^{2}$ \quad Jungyu Jin$^{3}$ \quad Younghyun Cho$^{3}$ \quad  Eunhyeok Park$^{3}$ \\
$^{1}$Department of Convergence IT Engineering \\
$^{2}$Department of Computer Science and Engineering \\
$^{3}$Graduate School of Artificial Intelligence \\
Pohang University of Science and Technology (POSTECH) \\
\texttt{\{changhun.lee, minsang5749, jgjin0317, yhcho97, eh.park\}@postech.ac.kr} \\
}

\begin{document}
\maketitle
\begin{abstract}
While many advanced LLMs are designed to handle long sequence data, we can still observe notable quality degradation even within the sequence limit. In this work, we introduce a novel approach called Scaling to Emphasize Attention for Long-context retrieval (SEAL), which enhances the retrieval performance of large language models (LLMs) over long contexts. 
We observe that specific attention heads are closely tied to long-context retrieval, showing positive or negative correlation with retrieval scores, and adjusting the strength of these heads boosts the quality of LLMs in long context by a large margin. Built on this insight, we propose a learning-based mechanism that leverages generated data to emphasize these heads.
By applying SEAL, we achieve significant improvements in long-context retrieval performance across various tasks and models. Additionally, when combined with existing training-free context extension techniques, SEAL extends the contextual limits of LLMs while maintaining highly reliable outputs.
\end{abstract}

\section{Introduction}
Large Language Models (LLMs) (\citealp{brown2020language}; \citealp{radford2019language}; \citealp{touvron2023llama}), grounded in the highly effective self-attention mechanism of the Transformer architecture \citep{vaswani2017attention}, have demonstrated remarkable proficiency in capturing and modeling global dependencies within a given context. As LLMs continue to advance, there has been an increasing demand for their deployment in long-context applications, including document-level understanding, code synthesis, and multi-turn conversation.

However, this trend has also underscored a critical limitation of LLMs: the noticeable degradation in output quality when processing longer context data, even within the predefined context window (\citealp{liu2024lost}; \citealp{he2024never}). Although state-of-the-art models are designed to accommodate extended contexts, recent studies (\citealp{li2023long}; \citealp{hsieh2024ruler}) have revealed that these models still suffer from significant performance degradation as the context length approaches its upper limit. For example, even in relatively straightforward tasks such as common word extraction, advanced LLMs capable of handling over 100K tokens have been observed to produce hallucinated outputs when processing inputs well below this threshold. This limitation, which is not evident when handling short to moderate context lengths, suggests that it does not stem from the model’s inherent knowledge capacity. Instead, it is more plausibly attributed to intrinsic biases, such as locality preferences induced by skewed datasets.

\begin{figure}[t]
\centering
\includegraphics[width=\linewidth]{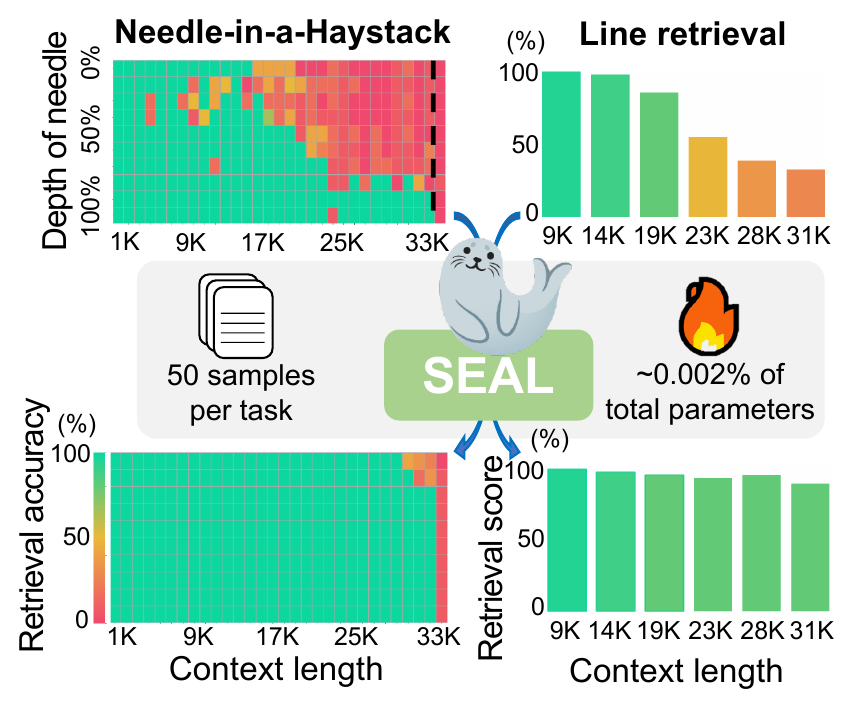}
\caption{Overview of the proposed SEAL and corresponding retrieval score improvements for LongChat-7B-v1.5-32K (\citealp{li2023long}) model.}
\vspace{-2mm}
\label{fig:overview}
\end{figure}

In this study, we aim to address this limitation of LLMs, with a particular focus on long-context retrieval tasks designed to generate answers through simple rule-based reasoning—such as recalling a specific word or counting occurrences—using synthetic data as a key benchmark, thereby minimizing the influence of parametric knowledge.
Our approach is grounded in a key insight: well-trained LLMs inherently possess the capability to infer information accurately regardless of context length, yet biases embedded in their trained parameters often result in performance degradation. For a representative long-context retrieval benchmark, we observed that some key attention components influence long-context retrieval quality, and adjusting their strength has a large impact on accuracy.

Building upon these observations, we propose a novel approach, \textit{Scaling to Emphasize Attention for Long-context Retrieval} (SEAL). SEAL is designed to enhance the attention scores of LLMs that are particularly crucial for long-context tasks by leveraging synthetic data formatted to align with the target task specifications. 
Specifically, head-wise and channel-wise learnable scales for attention output are fine-tuned for SEAL-H (head) and SEAL-C (channel), respectively. Through this process, SEAL not only probes the importance of each attention component but also adjusts the scaling to enhance retrieval performance.
With less than an hour of gradient-based fine-tuning to adjust scales for attention, SEAL significantly improves long-context retrieval quality, without incurring the computational overhead during inference.
Furthermore, SEAL can be applied in conjunction with training-free context length extension methods.
This approach effectively increases both the actual context window size and the effective context window size, which defines the range where output quality is maintained.
For more practical applications, we also provide a comprehensive analysis of the transferability of SEAL across different tasks.

\section{Related work}
\paragraph{Benchmarks for Long-Context LLMs}
Several benchmarks have been proposed to evaluate the retrieval and reasoning capabilities of long-context LLMs. Needle-in-a-Haystack (\citealp{gkamradt2023needle}) inserts a random fact or statement (`needle') into a long-context text (`haystack') and asks the model to retrieve the needle. 
LongEval (\citealp{li2023long}) line retrieval is the task of retrieving the corresponding digit given a key within a long text consisting of sentences with a line key and a value of up to five digits. RULER \cite{hsieh2024ruler} is a benchmark consisting of 13 tasks across 4 categories, designed to comprehensively assess long-context understanding capabilities based on synthetic examples.

\section{Motivation}
\label{sec:headwise_pruning}
In this section, we present our observation of retrieval inaccuracies in large language models (LLMs) as they approach their context length limit, along with the motivation to address this challenge. As illustrated in \Cref{fig:headwise_pruning}(c), the LongChat-7B-32K model exhibits significant performance degradation within the 19K–31K token range, despite being designed and trained to handle sequences of up to 32K tokens. This degradation is not limited to a single model but is also evident in other LLMs. Therefore, mitigating the decline in performance at long context lengths is essential to ensure the reliable utilization of LLMs in extended-context scenarios.

To address this challenge, we propose an optimistic hypothesis: \textbf{\textit{if we can identify and enhance the attention heads specialized in long-context retrieval, we may significantly improve LLM performance in this area.}} Previous research on Transformer-based architectures (\citealp{elhage2021mathematical}; \citealp{ferrando2024primer}) has demonstrated that attention heads—key components of these architectures—perform distinct roles such as copying, retrieval, and assessing relevance, collectively shaping the network's overall functionality. 
Notably, our observations indicate that certain attention heads are specialized in managing retrieval in long sequences. By selectively emphasizing and strengthening these heads, we aim to enhance the performance of LLMs in long-context scenarios.

\subsection{Attention Per-head Pruning}
\begin{figure}[t]
\centering
\includegraphics[width=\linewidth]{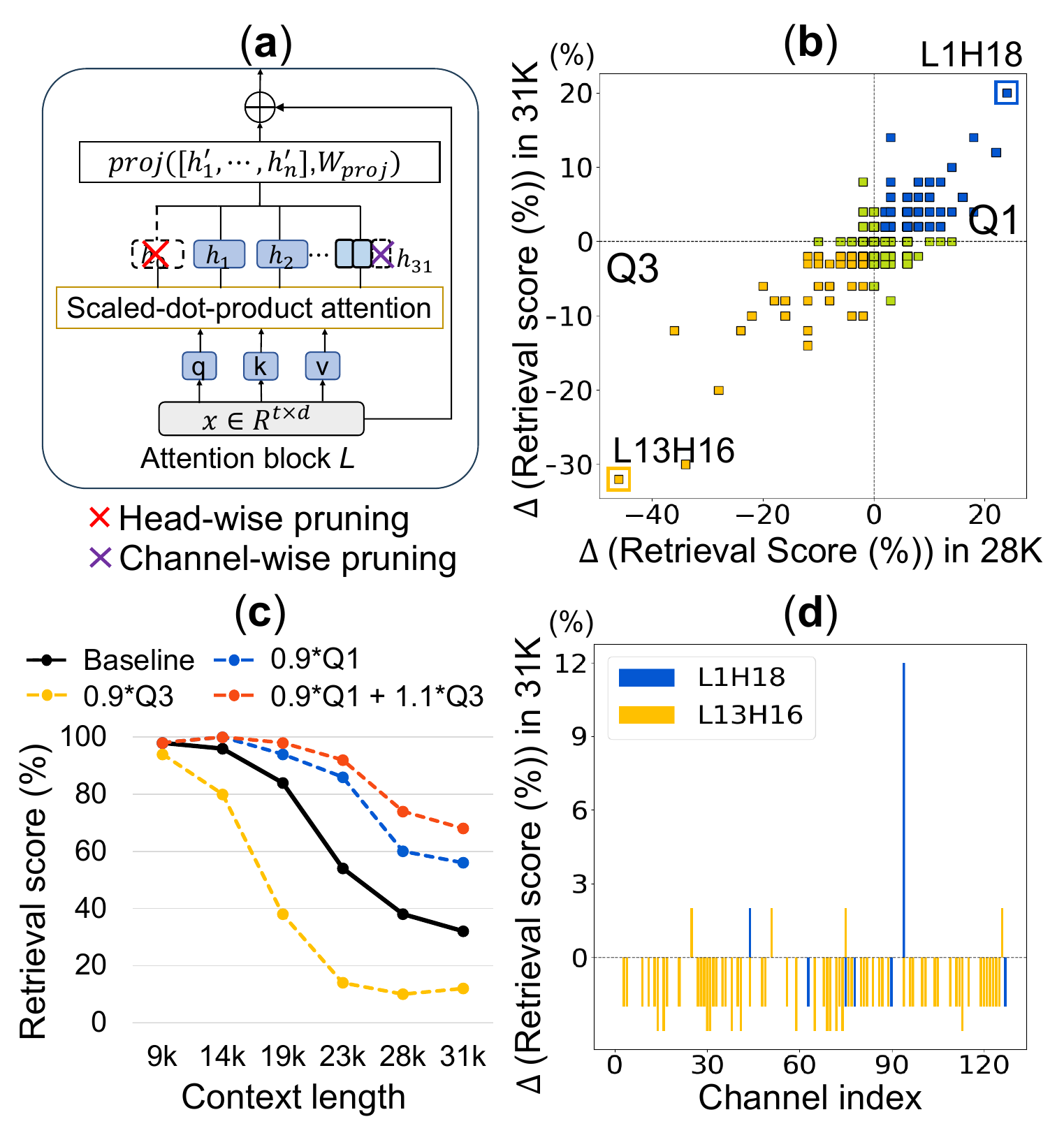}
\caption{Changes in retrieval scores (\%) with different settings. (a) Overview of pruning settings, (b) head-wise pruning results, (c) retrieval scores of scaling multiple heads, and (d) channel-wise pruning results. L$x$H$y$ refers to the $y$-th head of the $x$-th Transformer decoder block (zero-based indexing).}
\vspace{-2mm}
\label{fig:headwise_pruning}
\end{figure}

To validate this hypothesis, we first examined whether each attention head contributes differently to the retrieval process and sought to identify attention heads specialized for retrieval tasks. Our experimental design is straightforward: as illustrated in \Cref{fig:headwise_pruning}(a), we pruned one attention head at a time in the LongChat-7B-v1.5-32K model (\citealp{li2023long}) and compared the resulting accuracy to the original model’s accuracy. We employed a simple Line retrieval task from the LongEval benchmark (\citealp{li2023long}), where the goal is to retrieve randomly located up to five-digit numbers within a given text.

Interestingly, an important observation emerges: pruning certain attention heads can, in fact, improve retrieval performance. As illustrated in \Cref{fig:headwise_pruning}(b), \textbf{the impact of pruning each attention head varied significantly, with accuracy changes of approximately ±20\% or more.} Notably, these head-specific effects, both positive and negative, were consistently observed across mid-length (x-axis) and long-length (y-axis) contexts.

\subsection{Attention Head-wise Scaling}
\label{sec:head_wise_scaling}
Building on this observation, we sought to investigate whether the effects of head pruning could be combined to achieve more generalized performance improvements. To validate this hypothesis, \textbf{we adjusted the scaling of the multiple identified heads together to assess whether holistic accuracy gains could be achieved.} 
First, we divided the quadrants in \Cref{fig:headwise_pruning}(b) based on baseline performance (change in accuracy: 0) along the x- and y-axes. Scaling down the influence of all heads in the first quadrant (Q1)—where pruning positively impacted retrieval—by a factor of 0.9 improved accuracy from 32\% to 56\% at an input length of 31K tokens (blue dotted line in \Cref{fig:headwise_pruning}(c)). 
In contrast, scaling down the heads in Q3—where pruning degraded retrieval performance—by 0.9 led to a significant accuracy drop (yellow line in \Cref{fig:headwise_pruning}(c)). Notably, scaling Q1 heads by 0.9 while emphasizing Q3 heads by 1.1 simultaneously resulted in an even greater improvement in performance (red line). This suggests that jointly controlling the influence of these heads can significantly enhance retrieval accuracy.

\subsection{Attention Channel-wise Scaling}
\label{sec:channel_wise_pruning}
As highlighted in Quantizable Transformers (\citealp{bondarenko2023quantizable}), prior research suggests that specific channels within Transformer models manage syntactic elements such as delimiter tokens and even encode task-specific knowledge (\citealp{rudman2023outlier}). 
Building on this insight, we further conducted channel-wise pruning experiments on the LongChat-7B-v1.5-32K model (\citealp{li2023long}), where the hidden dimension of multi-head attention is 4096, which consists of 32 heads, each with a channel dimension of 128 ($32 \times128=4096$).

In our previous head-wise pruning experiment, the most performance-improving head (Layer 1, 18th head: L1H18) was identified (\Cref{fig:headwise_pruning}(b)). We then conducted a channel-wise pruning experiment by sequentially pruning each of the 128 channels within this L1H18 head, one channel at a time.

The results in \Cref{fig:headwise_pruning}(d) show that pruning the 94th channel of L1H18 head led to most of the performance improvement, 12\%, whereas pruning other channels within L1H18 head sometimes even resulted in performance degradation. We also conducted the same channel-wise pruning experiment on the head that caused the most significant performance drop in head-wise pruning (Layer 13, 16th head: L13H16) and observed similar variations in the impact of different channels on long-context retrieval.
\textbf{These findings underscore the importance of channel-wise attention manipulation beyond head-level adjustments.}

\begin{figure*}
\centering
\includegraphics[width=\linewidth]{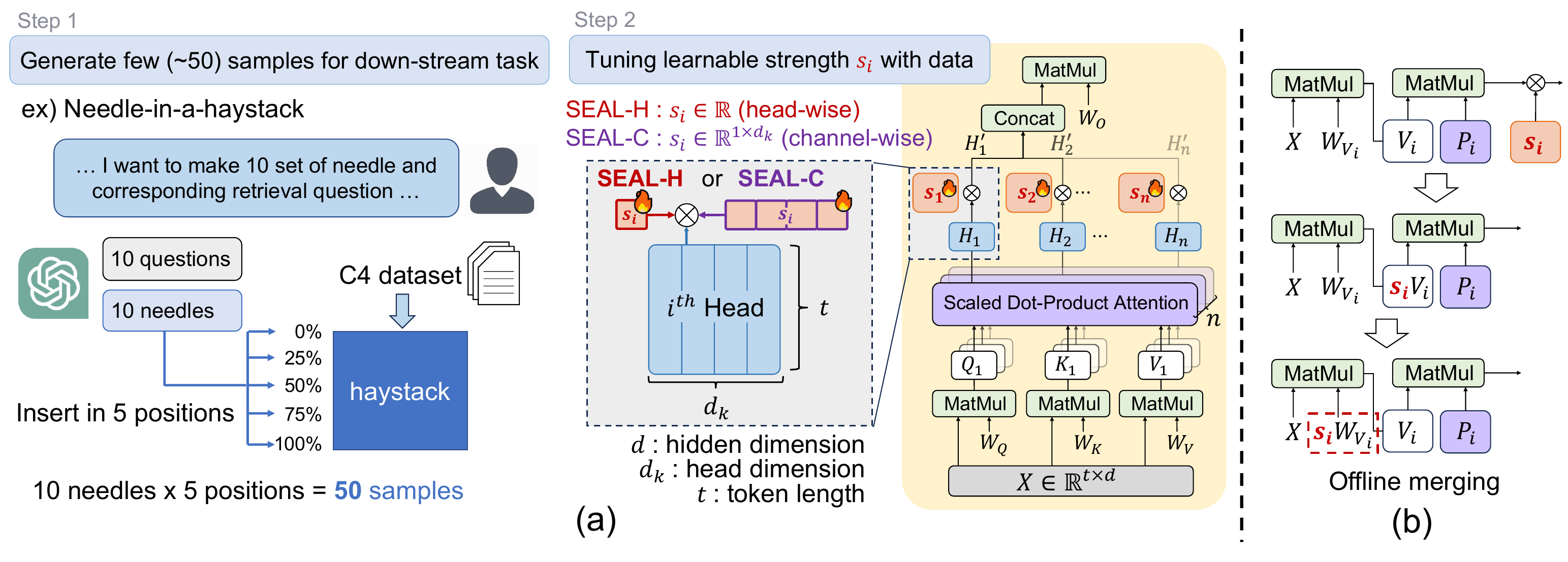}
\caption{(a) The overview of the proposed SEAL method. SEAL-H (head) or SEAL-C (channel) can be used depending on scaling granularity. (b) Offline merging process of the learned scale.}
\label{fig:seal_overview}
\end{figure*}

\section{Proposed method: SEAL}
\label{sec:seal}
Building on these invaluable observations, we introduce a novel method called Scaling to Emphasize Attention for Long-Context Retrieval (SEAL), a framework designed to enhance the long-context retrieval performance of existing LLMs. SEAL updates the attention strength of LLMs without altering their learned behavior. Based on the observations in \Cref{sec:head_wise_scaling} and \ref{sec:channel_wise_pruning}, SEAL aims to modulate the strength of each attention head or channel. 
Since sequentially performing head or channel-wise pruning to identify the influence of all attention components for each task is infeasible, we adopt a data-driven, first-order optimization approach instead.
\Cref{fig:seal_overview}(a) provides an overview of SEAL. The methodological contributions of SEAL lie in two key areas: \textbf{(1) format-aware generation of training datasets} and \textbf{(2) the design of an appropriate learnable space tailored to enhance retrieval performance.}

\subsection{Format-aware data synthesis}
During the dataset generation stage, we observed that SEAL does not focus on the inherent value of real-world data, but rather on the format and representation of target long-context tasks. For model updates, we generated synthetic training data using the task domain’s format instead of real data with meaningful values, and used this synthetic data to modulate attention strength. 
We generated 50 sample input-output pairs for the given downstream long-context task. To prevent data contamination, we ensured that only the format remained consistent while generating random content.
The method for obtaining format-specific samples may vary depending on the downstream task. The left side of \Cref{fig:seal_overview}(a) illustrates the pipeline for generating training samples for the Needle-in-a-Haystack task (\citealp{gkamradt2023needle}) using an LLM, and below are example samples created for (a) line retrieval and (b) Needle-in-a-Haystack tasks.

\tcolorbox[boxsep=1mm,colback=gray!10!white, colframe=black!50!black,top=3pt,bottom=3pt,left=2mm, right=2mm]
\small{
(a) \textbf{Prompt:} ... line righteous-ethernet: REGISTER\_CONTENT is $<$40779$>$ ...

    \textbf{Answer\_string:} The $<$REGISTER\_CONTENT$>$ in line righteous-ethernet is 40779. \\

(b) \textbf{Prompt:} ... Based on the content of the book, Question: What is immediately noticeable upon entering the room?

    \textbf{Answer\_string:} Immediately noticeable upon entering the room is the large oak table positioned beneath the chandelier.
    }
\endtcolorbox

\subsection{Learnable space design: SEAL-H and SEAL-C}
Using the generated data, we update a learnable scaling mechanism for attention components. Inspired by the insights from pruning experiments, we propose two granularities for attention control. The first is SEAL-H (head), which applies a learnable scalar to each attention head (\Cref{fig:seal_overview}(a)). This approach enables us to assess the influence of individual heads on retrieval while jointly learning an appropriate scaling for long-context scenarios. The second is SEAL-C (channel), which employs a learnable vector for the hidden dimension of each attention output (\ie channel-wise scaling). Although SEAL-C requires more learnable parameters than SEAL-H, it offers finer-grained manipulation of attention head outputs, potentially leading to improved performance. Additional comparisons of SEAL-H and SEAL-C are provided in the \Cref{sec:appendix_seal_hc_comparison}.

\subsection{Practicality of SEAL: offline merging}
While SEAL-C and SEAL-H modulate attention strength at the channel and head levels, respectively, this modulation can be merged into the weights of adjacent layers, such as the \(v\_proj\) or \(o\_proj\) layers in Llama. For example, as illustrated in \Cref{fig:seal_overview}(b), the learned scaling factors can be multiplied offline along the output channel dimension of the \(v\_proj\) weights, ensuring \textbf{no additional computational overhead} during inference. For models utilizing Grouped-Query Attention \cite{ainslie2023gqa}, the scaling can instead be applied offline along the input channel dimension of the \(o\_proj\). Thus, while SEAL updates the attention strength, it introduces no inference-time overhead, underscoring the practicality of the SEAL framework.

\section{Experimental Results}

To validate the effectiveness of the proposed SEAL, we first evaluated its retrieval performance on long-context inputs for three widely-used benchmarks: line retrieval from LongEval, Needle-in-a-Haystack, and RULER, a complicated benchmark. 

\textbf{Models:} We validated SEAL on six models: LongChat-7B-v1.5-32K / Mistral-7B-Instruct-v0.2 (32K) (\citealp{jiang2023mistral}) / Vicuna-7B-v1.5-16K (\citealp{chiang2023vicuna}) / Vicuna-13B-v1.5-16K / LongChat-13B-16K / Llama-3.1-8B-Instruct model (128K) (\citealp{dubey2024llama}). We used the Llama-3.1-8B-Instruct model exclusively for RULER, since the model is known for its strong retrieval performance on relatively easy benchmarks.

\textbf{Settings:} We utilized the Axolotl\footnote{https://github.com/axolotl-ai-cloud/axolotl} framework to tune SEAL. The tuning was performed using the AdamW optimizer without learning rate decay, and all models were tuned for 1 epoch. 
Please refer to the appendix for details on the training configurations, such as learning rates.
A single A100 80GB GPU was used for both tuning and evaluation.

\textbf{Dataset generation:} We used 50 generated samples for each task. 
Models supporting 32K context window length were tuned with samples containing 31K input tokens, while models supporting 16K context window length used 16K input tokens.
For the 7B models, tuning with the 31K dataset took about 40 minutes, and tuning with the 16K dataset took about 10 minutes.

\label{sec:experiments}

\subsection{Qualitative Analysis with Circuit Analysis}

Before evaluating SEAL’s performance on downstream tasks, we first conducted a qualitative analysis to provide a deeper understanding of \textbf{how the proposed SEAL contributes to improving retrieval performance}. Recent studies have continuously sought to identify and interpret the internal mechanisms of LLMs and Transformers through analysis methods such as circuit analysis and logit attribution (\citealp{ferrando2024primer}; \citealp{lieberum2023does}). In this work, we employed the \textit{direct effect} (\citealp{lieberum2023does}), one of the most intuitive and effective approaches for circuit analysis.

Let $f(p)$ represent the hidden state output of each component (\textit{e.g.,} attention heads) for a prompt $p$ whose effect we aim to observe, and we denote the unembedding weight as $W_{unembed}$.  Then the direct effect without the normalization term can be expressed by the following equation:
\begin{equation}
  \Delta = W_{unembed} f(p)
\end{equation}

For the line retrieval task from the LongEval, we selected an example where the baseline LongChat-7B-v1.5-32K model produced an incorrect answer, while the tuned model with SEAL provided the correct answer. The selected example is shown below.

\tcolorbox[boxsep=1mm, colback=gray!10!white, colframe=black!50!black,
left=2mm, right=2mm]
\small{
\textbf{Prompt}: ...odd-shrimp: REGISTER\_CONTENT is $<$32616$>\backslash$nline verdant-efficiency: REGISTER\_CONTENT is $<$\textcolor{ForestGreen}{24819}$>\backslash$nline ...

\textbf{Question}: Tell me what is the \\ $<$REGISTER\_CONTENT$>$ in line verdant-efficiency? I need the number.\\

\textbf{Correct Answer}: The $<$REGISTER\_CONTENT$>$ in line verdant-efficiency is 248\textcolor{ForestGreen}{1}9.

\textbf{Wrong Answer}: The $<$REGISTER\_CONTENT$>$ in line verdant-efficiency is ``248\textcolor{red}{5}6".
}
\endtcolorbox
We analyzed the impact of each attention head on the final logit at the position of the last token in the input, just before the results diverged (\textcolor{ForestGreen}{1} and \textcolor{red}{5} in the example above), to examine the role SEAL played in the autoregressive generation process.

\begin{figure}[t]
\centering
\includegraphics[width=0.95\linewidth]{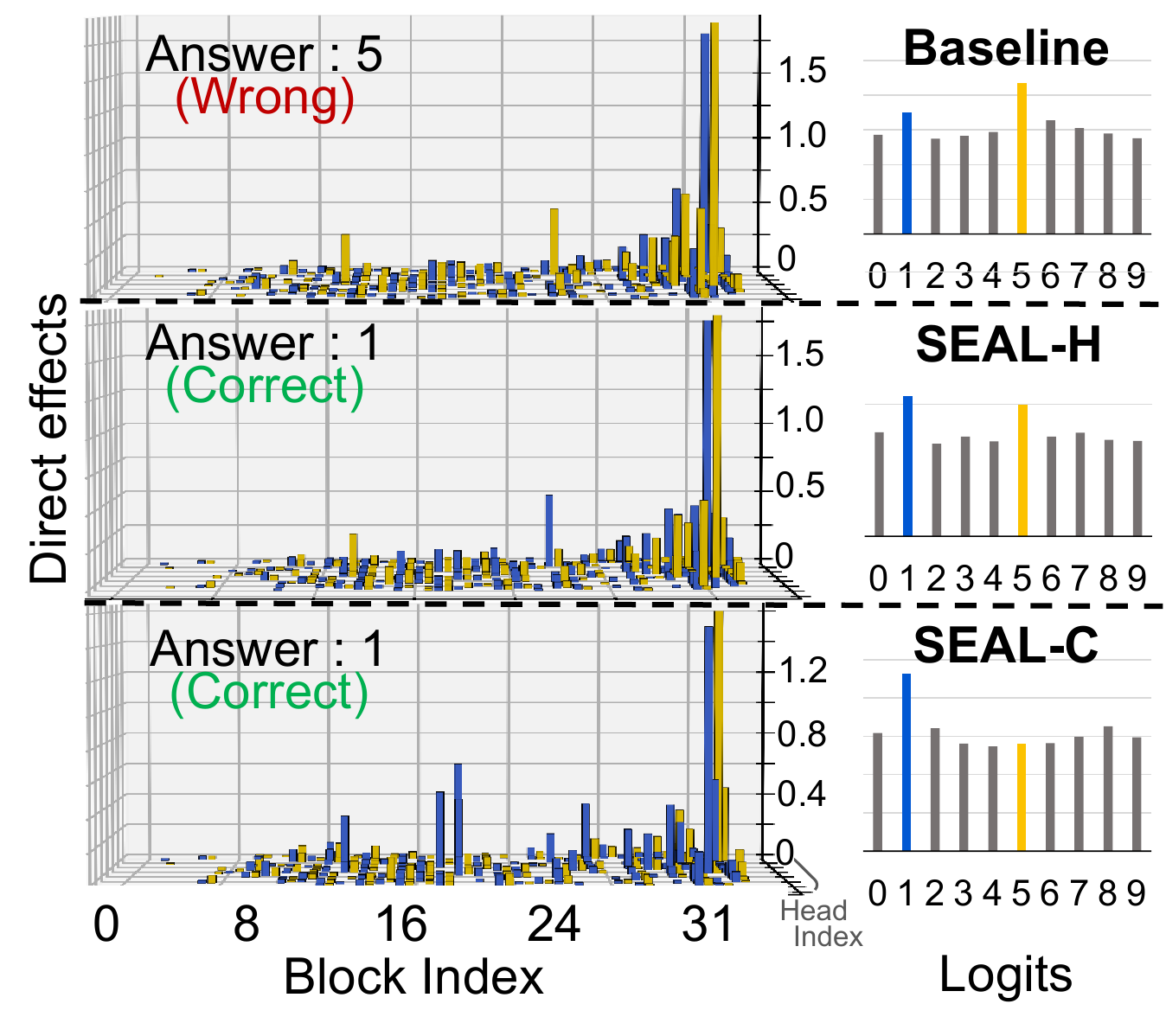}
\caption{Effects of attention heads on logits: Left: Direct effects of attention heads. Right: final logits before softmax function for each case.}
\label{fig:direct_effect}
\end{figure}

\Cref{fig:direct_effect} Left represents the direct effects of all attention heads, where the x- and y-axis represent the block and head index, respectively. 
The sum of the direct effects from all components for each token constitutes the final logits, and differences in this sum result in variations in model predictions. In the baseline model, certain attention heads exert a stronger direct effect on the incorrect digit \textcolor{red}{5}, leading to inaccurate predictions. However, the presence of a peak for the correct digit \textcolor{ForestGreen}{1} in the final logits suggests that the model does retain some internal retrieval capability for the correct answer.

In contrast, SEAL-H reduces the peak for \textcolor{red}{5} while increasing the peak for \textcolor{ForestGreen}{1} through appropriate head-wise scaling, ultimately improving retrieval by influencing the final logits. SEAL-C, utilizing channel-wise scaling, more precisely adjusts attention, ensuring that both the direct effect and the final logit value strongly favor \textcolor{ForestGreen}{1}. This demonstrates how SEAL effectively modifies retrieval outcomes, motivating further evaluation of its quantitative improvements across downstream retrieval tasks.

\renewcommand{\arraystretch}{0.9}
\begin{table}[t]
    \centering
    \setlength{\tabcolsep}{4pt}
    \caption{\label{tab:retrieval}
    Comparison of the line retrieval task scores.
    }
    \resizebox{\columnwidth}{!}{
        \begin{tabular}{cccccccc}
            \toprule[1.5pt]
            \textbf{Model} & \textbf{Method} & \textbf{9K} & \textbf{14K} & \textbf{19K} & \textbf{23K} & \textbf{28K} & \textbf{31K} \\
            \midrule[0.75pt] 
             \multirow{3}{*}{\shortstack{LongChat-7B\\-v1.5-32K}}& Baseline & 0.98 & 0.96 & 0.84 & 0.54 & 0.38 & 0.32  \\
              & \textbf{SEAL-H}  & 1.00 & 1.00 & 0.98 & 1.00 & 0.94 & 0.80 \\
              & \textbf{SEAL-C} &  0.98 & 0.96 & 0.94 & 0.92 & 0.94 & 0.88 \\
              \midrule[0.75pt] 
             \multirow{3}{*}{\shortstack{Mistral-7B\\-Instruct-v0.2}}& Baseline & 0.98 & 1.00 & 0.90 & 0.86 & 0.88 & 0.94 \\
              & \textbf{SEAL-H}  &  1.00 & 1.00 & 1.00 & 0.98 & 0.98 & 1.00 \\
              & \textbf{SEAL-C} &  1.00 & 1.00 & 1.00 & 1.00 & 1.00 & 0.98 \\
             \midrule[0.75pt] 
             \textbf{Model} & \textbf{Method} & \textbf{5K} & \textbf{7K} & \textbf{9K} & \textbf{12K} & \textbf{14K} & \textbf{16K} \\
             \midrule[0.75pt] 
             \multirow{3}{*}{\shortstack{Vicuna-7B\\-v1.5-16K}}& Baseline & 1.00 & 1.00 & 0.96 & 0.92 & 0.60 & 0.64 \\
              & \textbf{SEAL-H}  &  1.00 & 1.00 & 1.00 & 0.98 & 0.92 & 0.84 \\
              & \textbf{SEAL-C} &  1.00 & 1.00 & 1.00 & 0.94 & 0.96 & 0.98 \\
              \midrule[0.75pt] 
             \multirow{3}{*}{\shortstack{LongChat-13B\\-16K}} & Baseline & 0.96 & 0.94 & 0.92 & 0.92 & 0.80 & 0.60 \\
             & \textbf{SEAL-H} &  1.00 & 1.00 & 0.98 & 1.00 & 1.00 & 0.92 \\
             & \textbf{SEAL-C} & 1.00 & 1.00 & 1.00 & 1.00 & 1.00 & 0.96 \\
              \midrule[0.75pt] 
             \multirow{3}{*}{\shortstack{Vicuna-13B\\-v1.5-16K}}& Baseline & 0.98 & 0.98 & 0.94 & 0.88 & 0.68 & 0.42\\
             & \textbf{SEAL-H} & 1.00 & 1.00 & 0.96 & 1.00 & 0.96 & 0.94\\
             & \textbf{SEAL-C} & 1.00 & 1.00 & 0.96 & 0.98 & 0.98 & 0.94\\
            \bottomrule[1.5pt]        
        \end{tabular}
    }
\end{table}
\renewcommand{\arraystretch}{1.0}

\subsection{Results on line retrieval task}
\label{sec:line_retrieval}
In \Cref{tab:retrieval}, the baseline LongChat and Vicuna show significant score degradation as the input length increases. However, the proposed SEAL methods demonstrate dramatic improvements across all input lengths, with notable improvements for LongChat-7B (from 0.32 to 0.88) and Vicuna-13B (from 0.42 to 0.94). Mistral, while not experiencing a steep drop within the 32K length, also shows substantial improvements in almost all cases, reaching nearly 100\% accuracy when SEAL is applied.

These demonstrate that tuning the influence of attention is key to improving retrieval performance, a finding also validated through analysis. Additionally, SEAL-C generally exhibits higher performance, confirming that fine-grained control at the channel-wise level is important even within the influence of heads. 
When we validate LongChat-7B for the MMLU (\citealp{hendrycks2020measuring}) task, the results are 42.53 / 42.34 / 42.17 for baseline, SEAL-H, and SEAL-C, respectively. The MMLU scores remain nearly unchanged, indicating that SEAL effectively identifies and scales only the attention heads relevant to long-context retrieval. 

Note that the dataset contains only 50 samples, resulting in the use of fewer than 2 million tokens for adjusting attention intensity. This efficiency highlights how effectively SEAL identifies the core attention components for long-sequence retrieval.

\begin{figure}[t]
\centering
\includegraphics[width=\linewidth]{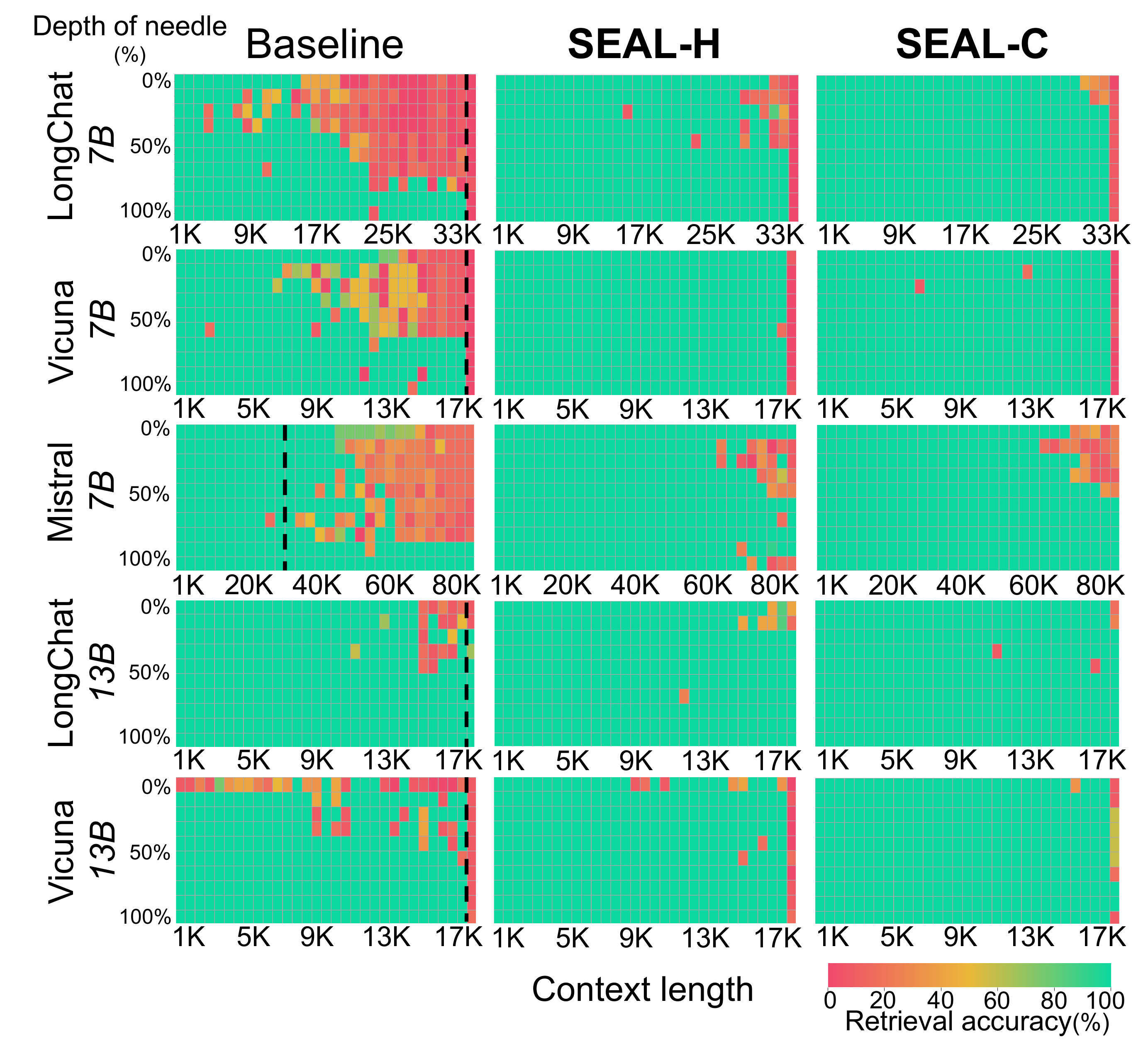}
\caption{Comparison of Needle-in-a-Haystack performances. The x-axis and y-axis represent the token length and the positions where the needle is inserted, respectively. The dotted black lines denote the context window limits of the original models.}
\label{fig:niah}
\end{figure}

\subsection{Results on Needle-in-a-Haystack task}

\Cref{fig:niah} presents the results of applying SEAL to the Needle-in-a-Haystack task. 
Despite using only 50 samples and training with synthesized needles that are different from the actual target needle, as depicted in \Cref{fig:seal_overview}(a), SEAL demonstrates remarkable performance improvement. 
Below is an example of correct and incorrect responses of the LongChat-7B-v1.5-32K model at a length of 20533 tokens, 22\% depth of needle insertion.
\tcolorbox[boxsep=1mm, colback=gray!10!white, colframe=black!50!black]
\small{
\textbf{Prompt}: ...It's a worrying prospect. The best thing to do in San Francisco is eat a sandwich and sit in Dolores Park on a sunny day. It would be a bummer to have another grim monoculture like...

\textbf{Question}: What is the best thing to do in San Francisco?\\

\textbf{SEAL-C (score: \textcolor{Green}{100\%})}: The best thing to do in San Francisco is eat a sandwich and sit in Dolores Park on a sunny day.

\textbf{Baseline (score: \textcolor{red}{8.3\%})}: Go to the top of the hill at Lands End and look out at the city.
}
\endtcolorbox

Although SEAL-H shows slightly lower performance than SEAL-C, it once again confirms that retrieval performance can be greatly recovered by simply adjusting the head-wise influence through scalar values, amounting to only 1024 parameters for the entire 7B model. Interestingly, in the case of Mistral, even though sample data were generated for a length of 31K for the SEAL, performance improved with inputs much longer than 31K.

\subsection{Results on RULER benchmark}
\label{sec:ruler}

\begin{table}[t]
    \centering
    \setlength{\tabcolsep}{4pt}
    \caption{\label{tab:ruler}
    Comparison of the RULER Benchmark scores. LongChat doesn't function properly for >32K inputs.
    }
    \renewcommand{\arraystretch}{0.85}
    \resizebox{\columnwidth}{!}{
        \begin{tabular}{c|ccccccc}
            \toprule[1.5pt]
            \textbf{Model} & \textbf{Task} & \textbf{Method} & \textbf{4K} & \textbf{8K} & \textbf{16K} & \textbf{32K} & \textbf{64K} \\
            \midrule[0.75pt]  
            \multirow{9}{*}{\shortstack{Llama-3.1\\-8B-Instruct}}  & \multirow{3}{*}{VT}     & Baseline & 99.3  & 98.8  & 99.3  & 97.7  & 94.3  \\
                &   & \textbf{SEAL-H}   & 99.8  & 99.7  & 99.8  & 99.8  & 98.1  \\
                &   & \textbf{SEAL-C}   & 100   & 100   & 100   & 99.4  & 99.1  \\
                \cmidrule(lr){2-8} & \multirow{3}{*}{CWE}      & Baseline & 99.5  & 94.3  & 53.9  & 2.6   & 0.1   \\
                &   & \textbf{SEAL-H}   & 100 & 99.8  & 98.7  & 95.8  & 21.8  \\
                &   & \textbf{SEAL-C}   & 100 & 99.6  & 99.5  & 99.7  & 95.7 \\
                \cmidrule(lr){2-8}
                & \multirow{3}{*}{FWE} & Baseline & 94.0  & 84.5  & 90.7  & 93.0  & 85.2  \\
                &   & \textbf{SEAL-H}   & 94.5  & 91.5  & 94.0  & 96.3  & 92.0  \\
                &   & \textbf{SEAL-C}   & 97.5  & 95.8  & 97.5  & 98.3  & 97.2  \\ 
            \midrule[0.75pt]  
            \multirow{9}{*}{\shortstack{Mistral-7B\\-Instruct-v0.2}} & \multirow{3}{*}{VT}     & Baseline & 99.6  & 95.9  & 88.1  & 86.5  & 70.8  \\
                &   & \textbf{SEAL-H}   & 99.5  & 99.0  & 97.1  & 94.9  & 80.8  \\
                &   & \textbf{SEAL-C}   & 99.5  & 99.4  & 98.8  & 98.4  & 85.8  \\
                \cmidrule(lr){2-8} & \multirow{3}{*}{CWE}      & Baseline & 98.7  & 90.4  & 69.8  & 28.5  & 0.3   \\
                &   & \textbf{SEAL-H}   & 97.8  & 94.3  & 86.4  & 66.4  & 59.8  \\
                &   & \textbf{SEAL-C}   & 98.8  & 99.0  & 98.0  & 96.9  & 83.6  \\
                \cmidrule(lr){2-8} & \multirow{3}{*}{FWE}      & Baseline & 85.3  & 77.3  & 94.3  & 91.8  & 77.5  \\
                &   & \textbf{SEAL-H}   & 90.3 & 86.3 & 98.0    & 93.7 & 80.3 \\
                &   & \textbf{SEAL-C}   & 99.0    & 94.3 & 96.8 & 98.3 & 91.8 \\
            \midrule[0.75pt] 
            \multirow{9}{*}{\shortstack{LongChat-7B\\-v1.5-32K}} & \multirow{3}{*}{VT}     & Baseline & 97.5  & 91.4  & 69.7  & 56.4  & -   \\
                &   & \textbf{SEAL-H}   & 97.8  & 97.4  & 93.6  & 69.8  & -   \\
                &   & \textbf{SEAL-C}   & 100   & 99.6  & 97.9  & 81.1  & -   \\
                \cmidrule(lr){2-8} & \multirow{3}{*}{CWE}      & Baseline & 73.7  & 39.6  & 23.9  & 37.7  & -   \\
                &   & \textbf{SEAL-H}   & 86.7  & 53.6 & 78.9  & 76.5 & -   \\
                &   & \textbf{SEAL-C}   & 88.2  & 71.0  & 97.0  & 93.7  & -   \\
                \cmidrule(lr){2-8} & \multirow{3}{*}{FWE} & Baseline & 59.2  & 72.7  & 50.2  & 69.2  & -   \\
                &   & \textbf{SEAL-H}   & 78.9  & 87.2  & 91.3  & 81.3  & -   \\
                &   & \textbf{SEAL-C}   & 92.2  & 92.7 & 96.0 & 94.5  & -   \\
            \bottomrule[1.5pt]
        \end{tabular}
    }
\end{table}

In the previous sections, we demonstrated that SEAL effectively boosts long-context retrieval scores across various model families and model sizes. To further investigate SEAL’s applicability, we evaluated whether SEAL (1) performs well in more complex long-context retrieval scenarios and (2) functions effectively for recent LLMs with extended context window sizes. To this end, we adopted the RULER (\citealp{hsieh2024ruler}) benchmark. Excluding overlapping categories from prior experiments, we selected three tasks: variable tracking (VT), common word extraction (CWE), and frequent word extraction (FWE). Although the RULER paper classified these tasks under categories other than retrieval, we consider them to be advanced forms of retrieval. For model selection, we included models with a context window size of 32K or more, including the recent Llama-3.1-8B-Instruct, which supports a 128K context window.

\Cref{tab:ruler} presents SEAL’s results across three subtasks, with each task evaluated on 200 samples. Notably, SEAL significantly enhances performance even for challenging long-context retrieval tasks that modern, sophisticated models struggle with. In particular, while the Llama-3.1-8B-Instruct model performs relatively well on the VT and FWE tasks, it exhibits a sharp performance drop in CWE as input length increases. Interestingly, SEAL substantially boosts retrieval performance, with SEAL-C achieving near-perfect scores. These results highlight that even advanced models with extended context window sizes still suffer from significant performance drops on difficult benchmarks for long-context retrieval. However, SEAL effectively mitigates this issue, demonstrating its robustness in improving long-context retrieval capabilities.

While our primary goal is to enhance the rule-based retrieval capabilities of LLMs using synthetic data, we additionally evaluated our approach on Document Question-Answering (QA) tasks from LongBench (\citealp{bai2024longbench}), a real-world long-context benchmark. The results and details are provided in \Cref{sec:appendix_longbench}.

\section{SEAL with context length extension}
There are two major challenges in handling long-context scenarios: (1) the gradual decline in performance within the context window, and (2) the length limitation of LLM's context window.

In this work, we address challenge (1) using the proposed SEAL.
However, our approach can be used in conjunction with existing methods that extend the context window length itself to address challenge (2). In fact, the validated LongChat from the previous sections is an example where the Llama (\citealp{touvron2023llama}) has already been extended with larger context windows via RoPE scaling and fine-tuning.
However, such tuning-based extensions come with significant costs in terms of time, data, and training infrastructure.

Therefore, to address challenge (2), training-free context length extension methods (\eg, NTK (\citealp{blocntkaware}), Self-Extend (\citealp{jin2024llm})) have 
gained attention but generally underperform compared to fine-tuning-based approaches (\eg, PI (\citealp{chen2023extending}), YaRN (\citealp{peng2023yarn})). 
Applying SEAL in conjunction with these training-free extension methods could combine the low training cost of SEAL with the benefits of tuning-free approaches, while also mitigating performance degradation.

\begin{figure}[t]
\centering
\includegraphics[width=0.90\linewidth]{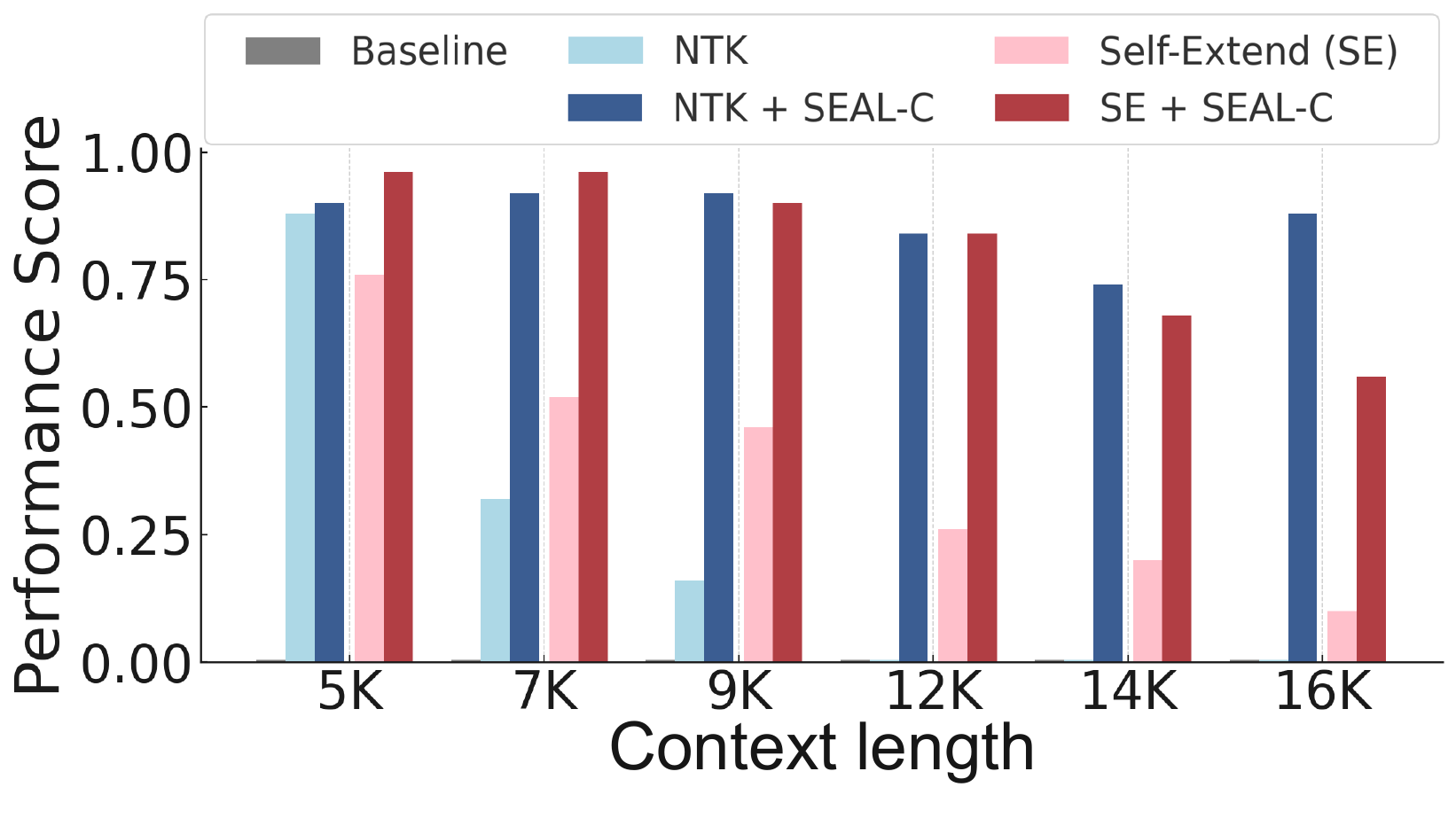}
\vspace{-2mm}
\caption{Line retrieval task scores for context length extension methods with and without SEAL in Llama-2-7b-Chat. Baseline gets zero scores for all lengths >4K. }
\label{fig:context_extension}
\end{figure}

\begin{figure}[t]
\centering
\includegraphics[width=\linewidth]{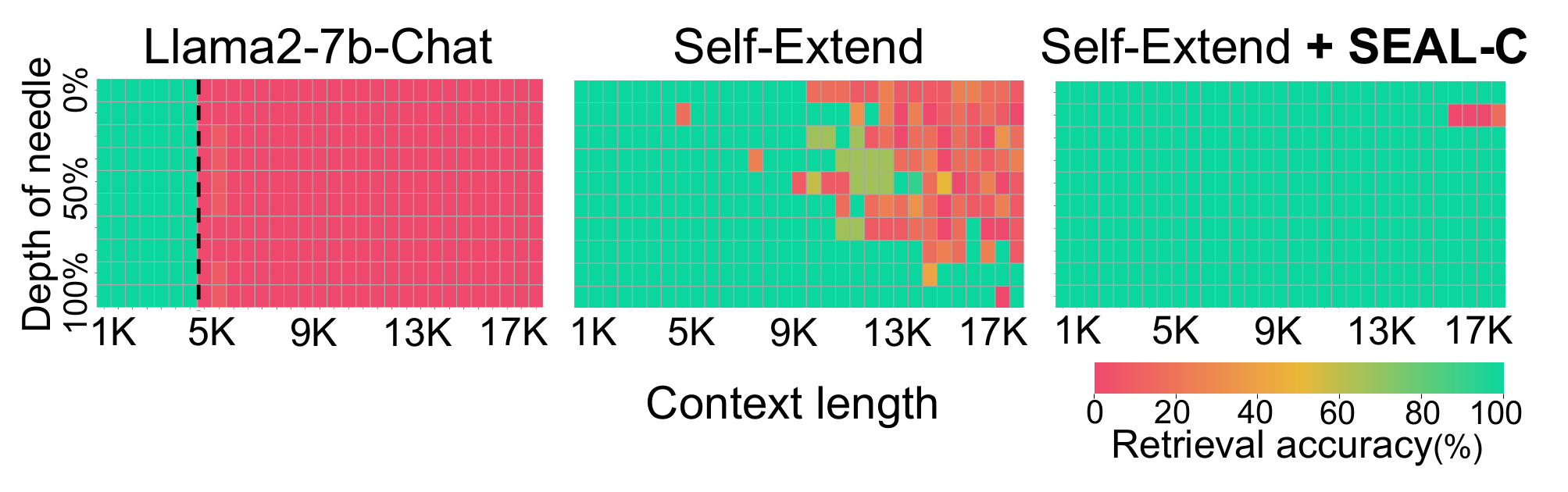}
\caption{The results of Needle-in-a-Haystack in Llama-2-7b-Chat. The dotted black line denotes the context window limits of the original Llama model: 4K tokens.}
\vspace{-2mm}
\label{fig:niah_extension}
\end{figure}

The results in \Cref{fig:context_extension} show that when extending the context length of Llama-2-7b-Chat (4K) to over 16K using only NTK or Self-Extend, the retrieval performance at lengths greater than 8K drops significantly. However, by utilizing SEAL in combination to adjust the attention influence, we can dramatically improve retrieval performance beyond the original base model's context window limitation (4K of Llama).
Notably, NTK is completely unable to retrieve information at lengths above 12K, yet with the application of SEAL, it achieves performance comparable to that at shorter input lengths.

Needle-in-a-Haystack task results in \Cref{fig:niah_extension} further demonstrate that SEAL significantly enhances the insufficient performance of the training-free context length extension methods. These results enable a practical approach to effectively increase the context length of the model at less than 1\% of the cost associated with fine-tuning-based context length extension methods by combining training-free context length extension with SEAL.

\section{Analysis on transferability of SEAL}
\label{sec:transferability}
\begin{figure}[t]
\centering
\includegraphics[width=0.85\linewidth]{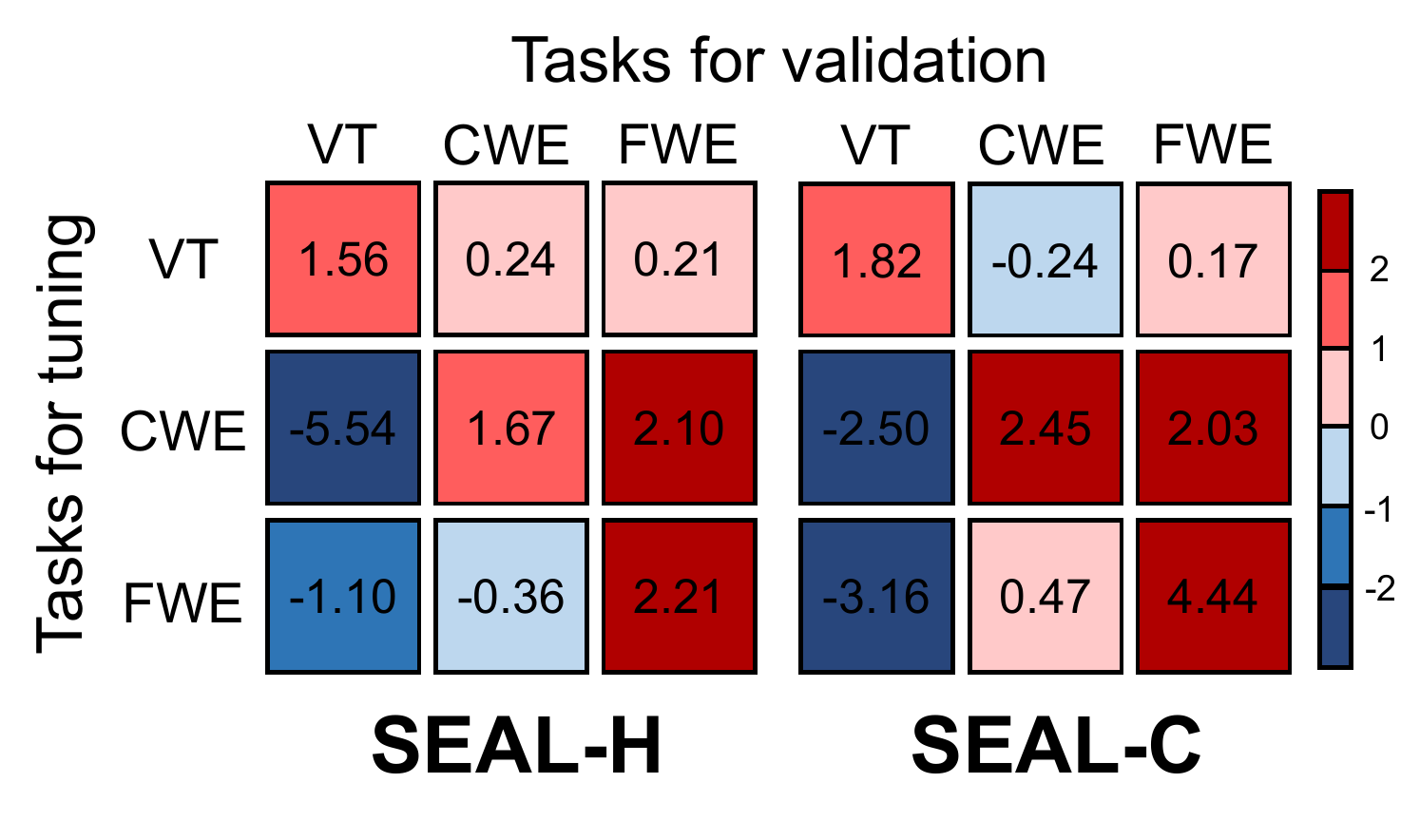}
\caption{The transferability metric of adapted attention scale of SEAL across RULER tasks.}
\label{fig:ruler_analysis}
\end{figure}

In this section, we aim to analyze whether the learned scale can transfer across tasks.
We first define a transferability metric from tuning task $A$ to validation task $B$ as follows: 
\begin{equation}
T_{A \rightarrow B} = \frac{\sum_{\ell} \left( S_B(\ell, \theta_A) - S_B(\ell, \theta) \right)}{\max_{\ell} S_B(\ell, \theta) - \min_{\ell} S_B(\ell, \theta)},
\end{equation}
where $\theta$ and $\theta_A$ denote the baseline model parameters and the model with scale tuned on task $A$, respectively. $S_B(\ell, \theta)$ denotes the validation score of task $B$ at length $\ell$, using $\theta$.
\Cref{fig:ruler_analysis} presents transferability across tasks from RULER. The y-axis represents the RULER tasks where SEAL is applied, while the x-axis indicates the tasks on which the learned scale was validated. 
This result provides some insights into the transferability of the learned scale across different retrieval tasks.

First, the red regions indicate performance improvement when the learned scale is applied to a validation task. Notably, scale transfer is effective between CWE and FWE, which belong to the same category.
Such transferability arises because SEAL leverages formatted data, allowing tasks with similar formats to benefit from cross-task adaptation. In contrast, scales tuned on CWE or FWE do not transfer well to VT.
These findings suggest that constructing a general long-context retrieval model may require gathering and tuning on a minimal set of representative samples from each task category, rather than every individual task. We leave this as a direction for future work. Finally, SEAL-C demonstrates higher transferability compared to SEAL-H.

\section{Comparison with In-Context Learning}
In-context learning (ICL) is a capability of LLMs to learn new tasks by leveraging pretrained contextual knowledge. ICL provides the task’s format information to the model through examples prepended to the prompt, similar to how SEAL fine-tunes using datasets focused on format information. Therefore, we compared the performance of ICL with SEAL.

We used 1-shot Prompting (\citealp{brown2020language}) with a 1K-length example and its answer, on the two models that showed the most significant performance drop even in the relatively simple Line retrieval task.
\Cref{tab:control_group} shows that one-shot prompting improves baseline performance for long inputs on the LongChat-7B model. However, it still demonstrates significantly lower performance compared to our SEAL, highlighting the effectiveness of SEAL in enhancing retrieval ability.
Additionally, the performance of ICL can sometimes be comparable to or even worse than the baseline due to the increased input length.

\renewcommand{\arraystretch}{0.9}
\begin{table}[t]
    \centering
    \setlength{\tabcolsep}{4pt}
    \caption{\label{tab:control_group}
    Comparison of the line retrieval task scores.
    }
    \resizebox{\columnwidth}{!}{
        \begin{tabular}{c|ccccccc}
            \toprule[1.5pt]
            \textbf{Model} & \textbf{Method} & \textbf{9K} & \textbf{14K} & \textbf{19K} & \textbf{23K} & \textbf{28K} & \textbf{30K} \\
            \midrule[0.75pt] 
             \multirow{8}{*}{\shortstack{LongChat-7B\\-v1.5-32K}}& Baseline & 0.98 & 0.96 & 0.84 & 0.54 & 0.38 & 0.32  \\
             \cmidrule{2-8}
             & One-shot  & 1.00 & 1.00 & 0.90 & 0.84 & 0.50 & 0.50 \\
             \cmidrule{2-8}
             & LoRA & 1.00 & 1.00 & 1.00 & 1.00 & 0.94 & 0.80\\
             & DoRA & 1.00 & 1.00 & 1.00 & 1.00 & 0.94 & 0.86\\
              \cmidrule{2-8}
              & \textbf{SEAL-H}  & 1.00 & 1.00 & 0.98 & 1.00 & 0.94 & 0.80 \\
              & \textbf{SEAL-C} &  0.98 & 0.96 & 0.94 & 0.92 & 0.94 & 0.88 \\
              \midrule[0.75pt] 
              \textbf{Model} & \textbf{Method} & \textbf{5K} & \textbf{7K} & \textbf{9K} & \textbf{12K} & \textbf{14K} & \textbf{15K} \\
            \midrule[0.75pt] 
             \multirow{8}{*}{\shortstack{Vicuna-13B\\-v1.5-16K}}& Baseline & 0.98 & 0.98 & 0.94 & 0.88 & 0.68 & 0.48\\
             \cmidrule{2-8}
             & One-shot  & 0.96 & 0.96 & 0.90 & 0.84 & 0.66 & 0.50 \\
             \cmidrule{2-8}
             & LoRA & 0.98 & 0.98 & 0.88 & 1.00 & 1.00 & 0.98 \\
             & DoRA & 1.00 & 0.98 & 0.90 & 1.00 & 1.00 & 0.98 \\
             \cmidrule{2-8}
             & \textbf{SEAL-H} & 1.00 & 1.00 & 0.96 & 1.00 & 0.96 & 1.00 \\
             & \textbf{SEAL-C} & 1.00 & 1.00 & 0.96 & 0.98 & 0.98 & 0.96 \\
            \bottomrule[1.5pt]        
        \end{tabular}
    }
\end{table}
\renewcommand{\arraystretch}{1.0}

\section{Comparison with Low-Rank Adaptation}
In SEAL, we focus solely on head-specific adaptation. However, the widely used Low-Rank Adaptation (LoRA) (\citealp{hulora}) technique encompasses a broader learning space that includes SEAL as a subset. For a more comprehensive analysis, we compare SEAL against LoRA and its recent variant, DoRA (\citealp{liu2024dora}). Both LoRA and DoRA with a low rank of \(r = 4\) were applied to all linear layers in the attention module (Q, K, V, and O) with a learning rate of 2e-4.

As shown in \Cref{tab:control_group}, SEAL achieves accuracy comparable to LoRA and DoRA on the line retrieval task. Notably, SEAL-H requires only \(L \times H\) learnable parameters (the number of blocks multiplied by the number of heads), amounting to just 1,024 parameters for the entire LongChat-7B model—making it highly parameter-efficient. Compared to LoRA, which updates all QKVO layers, SEAL-H attains similar performance while utilizing approximately 4,000 times fewer parameters. This demonstrates that SEAL is sufficient to enhance performance on these tasks effectively, while also validating our identification of key factors that contribute to improved retrieval performance.

\section{Conclusion}

The ability to retrieve and extract information from long-length input is an important component of the LLMs. Through our analysis, we found that there are attention components that have a good or bad impact on the retrieval scores. Based on this, we introduce SEAL, a novel attention strength scaling method to deliberately control the impact of each attention component. Despite using very few formatted sample data, SEAL significantly improves long-context retrieval performance. We believe that our insights will contribute to the advancement of long-context LLMs.

\section*{Limitations}
In this study, the training dataset used for SEAL included samples with a maximum length of 31K. If the length of each sample in the training dataset increases further, the GPU memory requirements for training will also grow, potentially necessitating the use of multi-GPU training. We expect that this issue can be mitigated by integrating SEAL with more efficient long-context training methods in future work. 

The only hyperparameter in SEAL is the learning rate. We observed that the optimal learning rate varies slightly depending on the difficulty or the characteristics of the downstream task, introducing a minor hyperparameter tuning requirement. However, since our target scenarios leverage synthetic data, they are free from the challenges of validation set construction. Additionally, tuning can be completed rapidly with as few as 50 samples, typically taking less than an hour. This ensures that the hyperparameter search incurs minimal overhead, making SEAL efficient in practical applications.

\section*{Acknowledgement}
This work was supported by Institute of Information \& communications Technology Planning \& Evaluation (IITP) and NRF grant funded by the Korea government(MSIT) (No. RS-2019-II191906, RS-2023-00213611, RS-2024-00415602, RS-2024-00457882)

\bibliography{seal_arxiv}

\newpage
\appendix

\section{Training Configurations}
\subsection{Learning rate}
The learning rate is the only hyperparameter in SEAL. In most cases, a learning rate of 1e-2 performed well; however, different values were used in some cases. The learning rates used to produce the results in this paper are listed in \Cref{tab:learning_rate1} and \Cref{tab:learning_rate2}.

\begin{table}[h]
    \centering
    \caption{Learning rates for the Line retrieval and Needle-in-a-Haystack tasks.}
    \resizebox{\linewidth}{!}{ 
    \begin{tabular}{c|c|c|c|c}
        \toprule
        & \multicolumn{2}{c|}{Line retrieval} & \multicolumn{2}{c}{Needle-in-a-Haystack}  \\
        \cline{2-5}
         & SEAL-H & SEAL-C & SEAL-H & SEAL-C\\
         \midrule
        LongChat-7B-v1.5-32K & \multirow{3}{*}{1e-2} & \multirow{5}{*}{2e-2} & \multirow{5}{*}{4e-2} & 5e-2 \\
        Mistral-7B-Instruct-v0.2 &  &  &  & 4e-2 \\
        \cline{5-5}
        Vicuna-7B-v1.5-16K &  &  &  & \multirow{3}{*}{5e-2}\\
        \cline{2-2}
        LongChat-13B-16K & \multirow{2}{*}{2e-2} &  &  & \\
        Vicuna-13B-v1.5-16K &  &  &  & \\
         \bottomrule
    \end{tabular}
    }
    \label{tab:learning_rate1}
\end{table}

\begin{table}[h]
    \centering
    \caption{Learning rates for the tasks of the RULER benchmark.}
    \resizebox{\linewidth}{!}{ 
    \begin{tabular}{c|c|c|c|c|c|c}
    \toprule
         & \multicolumn{2}{c|}{VT} & \multicolumn{2}{c|}{CWE} & \multicolumn{2}{c|}{FWE} \\
         \cline{2-7}
         & SEAL-H & -C & -H & -C & -H & -C \\
         \midrule
        Llama-3.1-8B-Instruct & \multirow{3}{*}{1e-2} & 1e-2 & \multirow{3}{*}{1e-2} & \multirow{3}{*}{1e-2} & \multirow{3}{*}{1e-2} & 5e-3 \\
        Mistral-7B-Instruct-v0.2 &  & 1e-2 & &  &  & 2e-2\\
        LongChat-7B-v1.5-32K &  & 2e-2 & & &  & 1e-2 \\
        \bottomrule
    \end{tabular}
    }
    \label{tab:learning_rate2}
\end{table}

\subsection{SEAL with Training-free context length extension}
For NTK, we set the scaling factor to 4 to extend the context length from 4096 to 16384. For Self-Extend, we set the group size to 6 and the neighbor window size to 1024, resulting in an extended context length of $(4096 - 1024) \times 6 = 18432$.

\section{Generating sample data for downstream task}
\label{sec:sample}
\subsection{Line retrieval}
LongEval provides generate\_testcases.py to create random data of the desired length. We created a prompt (input) for the sample utilizing that code.
The answer label for scale tuning is made as follows:
\tcolorbox[boxsep=1mm, colback=gray!10!white, colframe=black!50!black]
\small{
data['answer\_str'] = \\ f"The $<$REGISTER\_CONTENT$>$ in line \\ {data['random\_idx'][0]} is {data['expected\_number']}."
}
\endtcolorbox

We further used appropriate system prompts and conversation templates for each model when training with axolotl.

\subsection{Needle-in-a-Haystack}
The pipeline for generating sample data for Needle-in-a-Haystack is detailed in the \Cref{fig:seal_overview}(a).
We used the following input prompt to generate random needles using chatGPT\footnote{https://chatgpt.com}:
\tcolorbox[boxsep=1mm, colback=gray!10!white, colframe=black!50!black]
\small{
I am trying to test the retrieval performance of the model. I need needle sentences to find in a long context, with the corresponding retrieval question. 
Here is one example case: "needle": "The first thing you notice upon entering the room is the bright green chair sitting in the center facing the window.", "question": "What is the first thing you notice upon entering the room?".
I want to make 10 sets of needles and corresponding retrieval questions in jsonl format, like {"needle": "...", "question": "..."}.
Here are some restrictions about needles and questions.
1. Since the purpose is to test only retrieval performance, the less it is related to general knowledge, the better.
2. It is better to place the content corresponding to the question at the beginning of the needle sentence, like the given example.
3. Keep the length of the needle similar to or longer than the length of the example needle provided.
4. Please give variations to the format, "first thing".

Can you make 10 sets of examples for me?
}
\endtcolorbox

The 10 random needle and question pairs created from the above prompt are as follows: 
\tcolorbox[boxsep=1mm, colback=gray!10!white, colframe=black!50!black]
\small{
{"needle": "Immediately noticeable upon entering the room is the large oak table positioned beneath the chandelier.", "question": "What is immediately noticeable upon entering the room?"}

{"needle": "A striking feature of the room is the tall bookshelf that spans the entire length of the far wall.", "question": "What is a striking feature of the room?"}

{"needle": "Dominating the center of the room is a grand piano, its polished surface reflecting the light from the windows.", "question": "What dominates the center of the room?"}

{"needle": "Catching your eye as you step inside is the intricate tapestry hanging on the left wall, its colors vivid and bright.", "question": "What catches your eye as you step inside?"}

{"needle": "The first thing that draws your attention is the large framed photograph resting on the mantel.", "question": "What is the first thing that draws your attention?"}

{"needle": "Clearly visible as you enter is the large circular rug that covers most of the hardwood floor.", "question": "What is clearly visible as you enter?"}

{"needle": "What stands out immediately is the tall standing lamp positioned next to the armchair in the corner.", "question": "What stands out immediately in the room?"}

{"needle": "The most noticeable item upon stepping inside is the antique grandfather clock, ticking rhythmically in the corner.", "question": "What is the most noticeable item upon stepping inside?"}

{"needle": "Your attention is immediately drawn to the stained glass window, casting colorful patterns of light across the floor.", "question": "What is your attention immediately drawn to?"}

{"needle": "Visible as soon as you enter the room is a large painting of a landscape, mounted prominently on the main wall.", "question": "What is visible as soon as you enter the room?"}
}
\endtcolorbox

\subsection{RULER}
For the Mistral-7B-Instruct-v0.2 and LongChat-7B-v1.5-32K models, we generated 50 synthetic samples with random lengths between 8K and 31K.

For Llama-3.1-8B-Instruct, to accommodate the memory constraints of a single A100 GPU during training, we generated 50 samples with random lengths between 8K and 16K.

RULER benchmark provides code for each task to create random data of the desired length. We created a prompt (input) for the sample utilizing that code.

\section{Additional Comparisons of SEAL-H and SEAL-C}
\label{sec:appendix_seal_hc_comparison}
As explained in \Cref{sec:channel_wise_pruning}, which highlights the importance of channel-wise attention manipulation, SEAL-C (channel-wise) generally performs better in most use cases. Therefore, for practical applicability, we recommend using SEAL-C. However, even SEAL-H (head-wise), despite using only minimal learnable parameters (1/128 of SEAL-C and about 1/4000 of LoRA for LongChat-7B), already achieves significant performance improvements over the baseline. This again demonstrates that the attention head is a key component in enhancing retrieval performance, and channel-wise manipulation allows for more precise control over it.

Based on our experience, for in-depth analysis and understanding of LLM retrieval, it was highly beneficial to first identify the target attention head using SEAL-H and then perform a more fine-grained analysis using SEAL-C. Thus, a combined usage of both methods is effective for analysis.

As demonstrated in \Cref{sec:transferability}, SEAL-C also shows better transferability than SEAL-H.

\begin{table}[t]
\centering
\setlength{\tabcolsep}{4pt}
\caption{Comparison of scores on the LongBench Document QA tasks.}
\resizebox{\linewidth}{!}{
\begin{tabular}{lcccc|c}
\multicolumn{6}{c}{\textbf{Single-Doc QA}} \\
\toprule
 & MultiField & MultiField & Narrative & \multirow{2}{*}{Qasper} & \multirow{2}{*}{\textbf{Average}} \\
 & QA-EN & QA-ZH & QA & \\
\midrule
Baseline & 42.52 & 35.15 & 20.66 & 29.16 & \textbf{31.87} \\
\textbf{SEAL-H}   & 41.99 & 35.39 & 20.05 & 35.56 & \textbf{33.25} \\
\textbf{SEAL-C}   & 44.02 & 43.35 & 19.59 & 34.86 & \textbf{35.46} \\
\bottomrule
\end{tabular}
}

\vspace{1em}

\setlength{\tabcolsep}{3pt}
\resizebox{\linewidth}{!}{
\begin{tabular}{lcccc|c}
\multicolumn{6}{c}{\textbf{Multi-Doc QA}} \\
\toprule
 & HotpotQA & 2WikiMQA & MuSiQue & DuReader & \textbf{Average} \\
\midrule
Baseline & 33.12 & 23.89 & 14.49 & 21.66 & \textbf{23.29} \\
\textbf{SEAL-H}   & 38.77 & 23.92 & 18.58 & 22.92 & \textbf{26.05} \\
\textbf{SEAL-C}   & 45.13 & 32.50 & 22.93 & 24.52 & \textbf{31.27} \\
\bottomrule
\end{tabular}
}
\label{tab:doc_qa}
\end{table}

\section{Results on LongBench benchmark}
\label{sec:appendix_longbench}
Our primary focus is to improve the fundamental long-context retrieval capabilities of LLMs—tasks that often rely less on parametric knowledge and more on explicit rules or external formulas. This direction is essential, as rule-based reasoning serves as the foundation for a wide range of retrieval tasks, including those grounded in real-world data. Guided by this intuition, we intentionally enhance LLM performance using synthetic data, thereby decoupling SEAL’s contribution from improvements in the model’s parametric knowledge. Despite the apparent simplicity of these tasks, as shown in \Cref{sec:ruler}, even the latest Llama-3.1-8B-Instruct model struggles with basic rule-based operations, such as common word extraction (CWE). This observation highlights the pressing need to address challenges in synthetic retrieval tasks as a foundational priority.

On the other hand, we also provide additional results by extending our approach to real-world long-context tasks, especially LongBench (\citealp{bai2024longbench}), to demonstrate the practical applicability of SEAL beyond synthetic tasks.

Among the LongBench categories, Single-Document Question-Answering (QA) and Multi-Document QA require retrieving key passages across documents to properly answer the question. Accordingly, we measured the performance of our method on tasks categorized under Document QA. Since LongBench relies on real passages, we used 50 samples from the triviaqa\_e dataset, which belongs to the Few-Shot category and is not included in the document QA tasks used for evaluation, to prevent data contamination. Specifically, we sorted triviaqa\_e samples by context length in descending order and selected the top 50 samples as SEAL fine-tuning samples.
\Cref{tab:doc_qa} shows the improvements of LongBench's Document QA task scores with the proposed SEAL, on LongChat-7B-v1.5-32K model.

Despite the fact that Question-Answering heavily depends on other capabilities of LLMs, such as reasoning, applying SEAL led to performance improvements in most document QA tasks, significantly increasing the average score. This demonstrates that the retrieval performance gains achieved by SEAL also contribute to real-world long-context tasks like document QAs of LongBench. Interestingly, both SEAL-H and SEAL-C improved performance across all multi-doc QA tasks, particularly highlighting SEAL’s effectiveness in complex real-world tasks.

\section{Analysis on number of samples and learning rate}

\begin{table}[b]
    \centering
    \small
    \caption{Line retrieval results on LongChat-7B with a 31K input length, varying across learning rates (y-axis) and number of samples (x-axis).}
    \begin{tabular}{c|ccccc}
        \toprule
         & 10 & 30 & 50 & 70 & 99 \\
         \midrule
        5e-3 & 0.56 & 0.64 & 0.68 & 0.70 & 0.72 \\
        1e-2 & 0.68 & 0.74 & 0.78 & 0.82 & 0.82 \\
        2e-2 & 0.70 & 0.82 & 0.82 & 0.84 & 0.70 \\
        3e-2 & 0.76 & 0.84 & 0.90 & 0.86 & 0.82 \\
        \bottomrule
    \end{tabular}
    \label{tab:sweep}
\end{table}

\begin{figure}[h]
\centering
\includegraphics[width=0.75\linewidth]{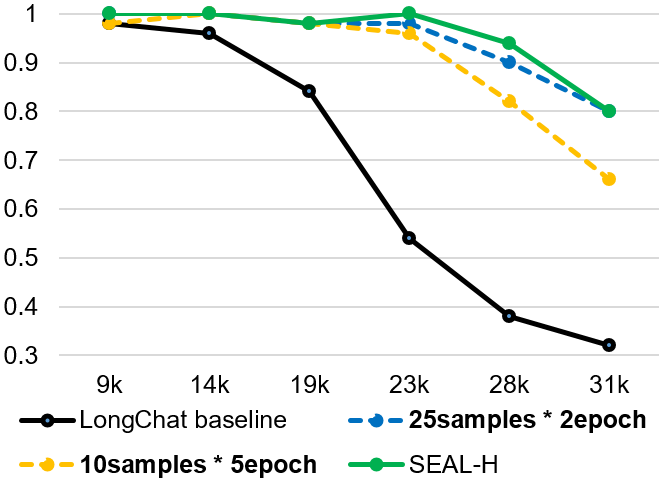}
\caption{Line retrieval results when using fewer samples than the default 50 samples.}
\label{fig:few_samples}
\end{figure}

One of the advantages of SEAL is that it can achieve significant performance improvements with a very small number of formatted data samples. 
To analyze the effects of both the number of samples and the learning rate on scale tuning, we tuned the scale of SEAL-H by sweeping over these two factors.
For this experiment, we generated a new set of 100 random samples for line retrieval using the same method proposed in \Cref{sec:sample}. The results of applying SEAL-H to LongChat-7B-v1.5-32K with different hyperparameter configurations are shown in \Cref{tab:sweep}. 
In general, we observed that performance improves as the number of samples increases, but found that around 50 samples are sufficient.

Additionally, we tested whether comparable performance improvements could be achieved using significantly fewer samples, with only 25 or 10 samples. In \Cref{fig:few_samples}, we compared tuning with 25 samples over 2 epochs and 10 samples over 5 epochs against the original SEAL-H (which used 50 samples over 1 epoch). The results show that even with as few as 25 samples, it is possible to achieve comparable performance. Although there is a relative performance decrease when tuning with only 10 samples for 5 epochs, it is remarkable that even with just 10 samples, there is a substantial improvement over the baseline. Preparing around 10 samples can be easily done by hand without the need for a complex data processing pipeline, which highlights the cost-effectiveness of the SEAL method.

\section{Analysis on parameters}
\Cref{tab:parameter_number} presents a comparison of the number of tunable parameters and their proportion relative to those used in the LoRA method. The results demonstrate that SEAL-H and SEAL-C utilize significantly fewer tunable parameters compared to low-rank adaptation methods, such as LoRA and DoRA.

\renewcommand{\arraystretch}{0.7}
\begin{table}[t]
    \centering
    \setlength{\tabcolsep}{4pt}
    \caption{\label{tab:parameter_number}
    Comparison of the number of tunable parameters and their ratio relative to those in the LoRA method. Here, K denotes $10^3$. 
    }
    \resizebox{\columnwidth}{!}{
        \begin{tabular}{ccrr}
            \toprule[1.5pt]
            \textbf{Model}  & \textbf{Method} & \multicolumn{1}{c}{\textbf{\# Params.}} & \multicolumn{1}{c}{\textbf{Ratio}} \\
            \midrule[1.25pt] 
            \multirow{5}{*}{LongChat-7B-v1.5-32K} & Baseline    & \multicolumn{1}{c}{-}    & \multicolumn{1}{c}{-} \\
    & \textbf{SEAL-H}          & \textbf{1.0K}       & \textbf{0.0002x} \\
    & \textbf{SEAL-C}          & \textbf{131.1K}     & \textbf{0.0313x} \\
    & LoRA            & 4194.3K    & 1.0000x \\
    & DoRA            & 4718.6K    & 1.1250x \\
            \midrule[0.75pt] 
            \multirow{5}{*}{Mistral-7B-Instruct-v0.2}  & Baseline    & \multicolumn{1}{c}{-}    & \multicolumn{1}{c}{-} \\
    & \textbf{SEAL-H}          & \textbf{1.0K}      & \textbf{0.0003x} \\
    & \textbf{SEAL-C}          & \textbf{131.1K}    & \textbf{0.0385x} \\
    & LoRA            & 3407.9K   & 1.0000x \\
    & DoRA            & 3735.6K   & 1.0962x \\
            \midrule[0.75pt] 
            \multirow{5}{*}{Vicuna-7B-v1.5-16K}        & Baseline    & \multicolumn{1}{c}{-}    & \multicolumn{1}{c}{-} \\
    & \textbf{SEAL-H}          & \textbf{1.0K}            & \textbf{0.0002x} \\
    & \textbf{SEAL-C}          & \textbf{131.1K}          & \textbf{0.0313x} \\
    & LoRA            & 4194.3K         & 1.0000x \\
    & DoRA            & 4718.6K         & 1.1250x \\
            \midrule[0.75pt]
            \multirow{5}{*}{LongChat-13B-16K}          & Baseline    & \multicolumn{1}{c}{-}    & \multicolumn{1}{c}{-} \\
    & \textbf{SEAL-H}          & \textbf{1.6K}            & \textbf{0.0002x} \\
    & \textbf{SEAL-C}          & \textbf{204.8K}          & \textbf{0.0313x} \\
    & LoRA            & 6553.6K         & 1.0000x \\
    & DoRA            & 7372.8K         & 1.1250x \\
            \midrule[0.75pt]
            \multirow{5}{*}{Vicuna-13B-v1.5-16K}       & Baseline    & \multicolumn{1}{c}{-}    & \multicolumn{1}{c}{-} \\
    & \textbf{SEAL-H}          & \textbf{1.6K}            & \textbf{0.0002x} \\
    & \textbf{SEAL-C}          & \textbf{204.8K}          & \textbf{0.0313x} \\
    & LoRA            & 6553.6K         & 1.0000x \\
    & DoRA            & 7372.8K         & 1.1250x \\
            \midrule[0.75pt]
            \multirow{5}{*}{Llama-3.1-8B-Instruct}     & Baseline    & \multicolumn{1}{c}{-}    & \multicolumn{1}{c}{-} \\
    & \textbf{SEAL-H}          & \textbf{1.0K}      & \textbf{0.0003x} \\
    & \textbf{SEAL-C}          & \textbf{131.1K}    & \textbf{0.0385x} \\
    & LoRA            & 3407.9K   & 1.0000x \\
    & DoRA            & 3735.6K   & 1.0962x \\
            \bottomrule[1.5pt]   
        \end{tabular}
    }
\end{table}

\section{Comparison with Low-Rank Adaptation: Needle-in-a-Haystack task}
\begin{figure}[t]
\centering
\includegraphics[width=\linewidth]{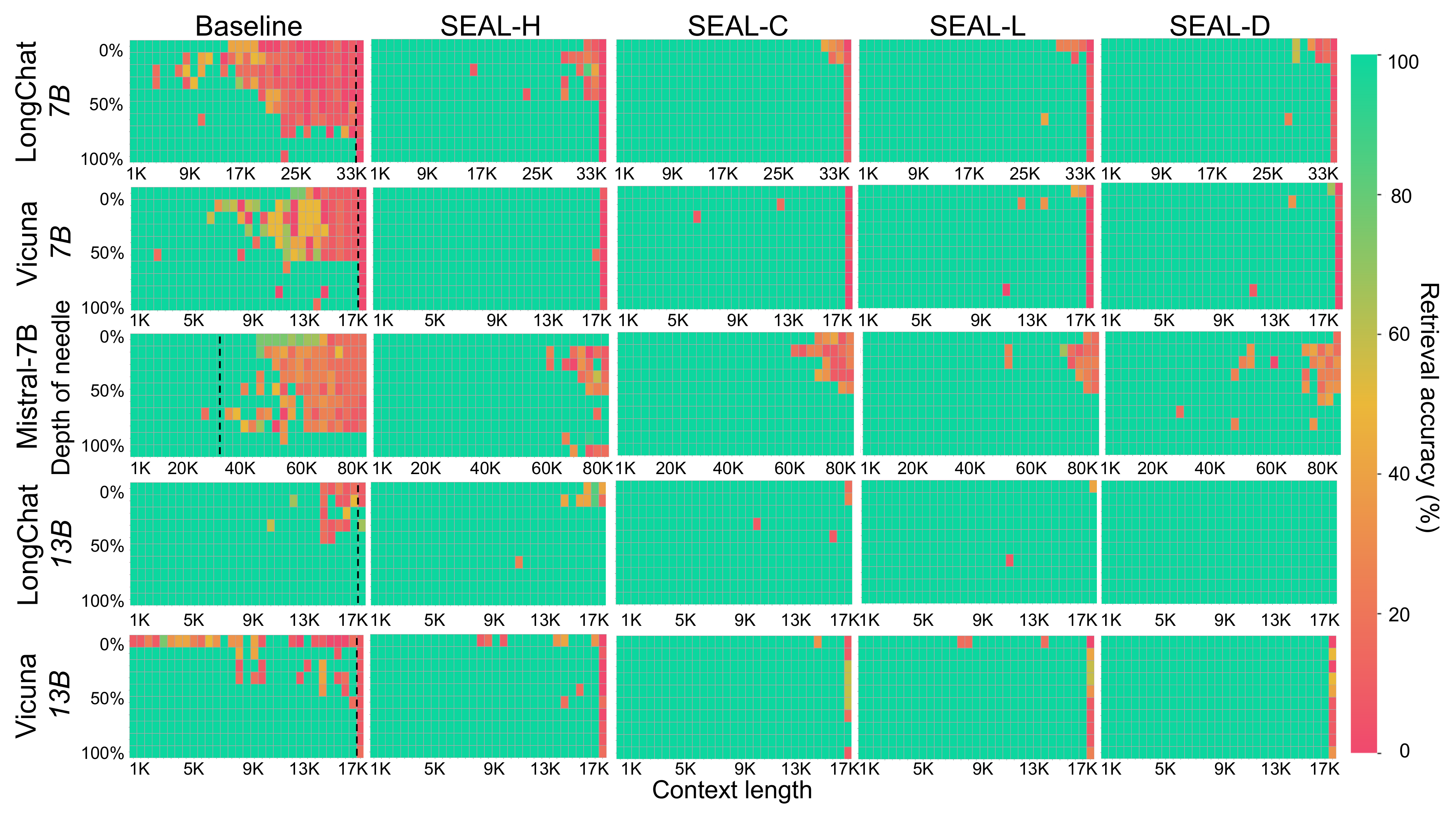}
\caption{Comparison of Needle-in-a-Haystack performance with Low-Rank Adaptation methods.}
\label{fig:niah_full}
\end{figure}

Additionally, we provide a comparison with low-rank adaptation methods for the Needle-in-a-Haystack task. \Cref{fig:niah_full} presents the results of applying SEAL, LoRA, and DoRA to the Needle-in-a-Haystack task. SEAL achieves performance comparable to low-rank adaptation methods in the Needle-in-a-Haystack task while utilizing significantly smaller learnable space.

\end{document}

%% file: math_commands.tex
\usepackage{amsmath,amsfonts,bm}

\def\eqref#1{equation~\ref{#1}}

\def\1{\bm{1}}

\DeclareMathAlphabet{\mathsfit}{\encodingdefault}{\sfdefault}{m}{sl}
\SetMathAlphabet{\mathsfit}{bold}{\encodingdefault}{\sfdefault}{bx}{n}

